\documentclass[conference]{IEEEtran}
\IEEEoverridecommandlockouts

\usepackage{amssymb}
\usepackage[cmex10]{amsmath}
\usepackage{stfloats}
\usepackage{graphicx}
\usepackage{subfigure}
\usepackage{tabularx}
\usepackage{epsfig,epsf,color,balance,cite}
\usepackage{verbatim}
\usepackage{url}
\usepackage{bm}

\usepackage{algorithm}
\usepackage{algorithmic}

\usepackage{color}
\definecolor{myc1}{rgb}{0,0,0}

\begin{document}

\title{Semantic Information Extraction for Text Data with Probability Graph}

\author{
\IEEEauthorblockN{Zhouxiang Zhao\IEEEauthorrefmark{1},
                  Zhaohui Yang\IEEEauthorrefmark{1}\IEEEauthorrefmark{2}\IEEEauthorrefmark{3},
                  Ye Hu\IEEEauthorrefmark{4},
                  Licheng Lin\IEEEauthorrefmark{1},
                 and Zhaoyang Zhang\IEEEauthorrefmark{1}\IEEEauthorrefmark{3}
                  }
	\IEEEauthorblockA{
			$\IEEEauthorrefmark{1}$College of Information Science and Electronic Engineering, Zhejiang University, Hangzhou, China\\
   			$\IEEEauthorrefmark{2}$ Zhejiang Lab, Hangzhou, China\\
			$\IEEEauthorrefmark{3}$Zhejiang Provincial Key Laboratory of Info. Proc., Commun. \& Netw. (IPCAN), Hangzhou, China
			\\
   $\IEEEauthorrefmark{4}$Department of Industrial and System Engineering, University of Miami, Coral Gables, USA\\
			E-mails: zhouxiangzhao@zju.edu.cn,
   yang\_zhaohui@zju.edu.cn,
   yxh1096@miami.edu,\\
   linlicheng@zju.edu.cn,
   ning\_ming@zju.edu.cn
		}
\thanks{This work is supported by Zhejiang Lab Program under grant K2023QA0AL02, Zhejiang Science and Technology Program under grant 2023C01021, National Natural Science Foundation of China under Grants U20A20158, and 61725104.}
\vspace{-3em}
}

\maketitle

\begin{abstract}
In this paper, the problem of semantic information extraction for resource constrained text data transmission is studied. In the considered model, a sequence of text data need to be transmitted within a communication resource-constrained network, which only allows limited data transmission. Thus, at the transmitter, the original text data is extracted with natural language processing techniques. Then, the extracted semantic information is captured in a knowledge graph. An additional probability dimension is introduced in this graph to capture the importance of each information. This semantic information extraction problem is posed as an optimization
framework whose goal is to extract most important semantic information for transmission. To find
an optimal solution for this problem, a Floyd's algorithm based solution coupled with an efficient sorting mechanism is proposed.
Numerical results testify the effectiveness of the proposed algorithm with regards to two novel performance metrics including semantic uncertainty and semantic similarity.

\end{abstract}

\IEEEpeerreviewmaketitle

\section{Introduction}
Over the past few decades, the development of mobile communication technology has greatly contributed to the progress of human society. In the 1940s, Shannon proposed information theory \cite{shannon1948mathematical}, which focused on quantifying the maximum data transmission rate that a communication channel could support. Guided by this fundamental theory, most existing communication systems have been designed based on metrics which concentrate on transmission rate. With the rapid increase in demand for intelligent applications of wireless communication, the future communication network will change from a traditional architecture that simply pursues high transmission rate to a new architecture that is oriented to complete tasks efficiently. There will be more and more tasks require low latency and high efficiency in future mobile information networks, which poses a huge challenge \cite{gunduz2022beyond,2023big} to the existing mobile communication systems. This trend gives rise to a new communication paradigm, called \emph{semantic communication}. Shannon's theory did not give importance to the semantic information of the data. It was considered to be largely irrelevant to communication. The concept of semantic communication was introduced by Shannon's collaborator Warren Weaver. Semantic communication is a new architecture that can integrate user needs and information meaning into the communication process. Traditional communication systems are primarily dedicated to reliably transmitting bit streams. It does not know the meaning of the message or what the goal of the message exchange is. Semantic communication systems, on the other hand, are dedicated to conveying the meaning in the message \cite{chaccour2022less,9832831} and can significantly reduce the required transmission channel bandwidth \cite{han2022semantic}. Semantic communication is expected to be a key technology for the future 6G mobile communication systems \cite{10024766,peng2022robust}.

 Recently, several works studied a number of problems related to semantic communications. The authors in \cite{xie2021deep} proposed a deep learning based semantic communication system for text transmission. To measure the performance of semantic communication, the authors also designed a new metric called sentence similarity. In \cite{weng2021semantic}, a deep learning-enabled semantic communication system for speech signals was designed. In order to improve the recovery accuracy of speech signals, especially for the essential information, it was developed based on an attention mechanism by utilizing a squeeze-and-excitation network. The work in \cite{tong2021federated} enabled the transmission of audio semantic information which captures the contextual features of audio signals. To extract the semantic information from audio signals, a wave to vector architecture based autoencoder that consists of convolutional neural networks was proposed. The authors in \cite{xie2020lite} proposed a lite distributed semantic communication system based on deep learning, named L-DeepSC, for text transmission with low complexity, where the data transmission from the IoT devices to the cloud/edge works at the semantic level to improve transmission efficiency. However, these existing works in \cite{xie2021deep,weng2021semantic,tong2021federated,xie2020lite} did not consider a flexible design that has the ability to adapt the content of the transmission in accordance with the communication environment.

The main contribution of this paper is a novel probability graph based important information extraction method, which enables the selection of the most important semantic information with in resource constrained networks. The key contributions are listed as follows:

\begin{itemize}
\item We propose a novel framework to extract important information for semantic text data transmission under communication resource constrains. In particular, we capture the semantic information in a knowledge graph, and introduce an additional relation probability dimension in the graph to capture the importance of the information. We formulate this information extraction problem in an optimization framework, and seek to find the
optimal strategy that finds the most important information to extract and transmit. 

\item To guarantee the effectiveness of the proposed semantic information extraction solution, we introduce two evaluation metrics: semantic uncertainty, which measure the clarity of the semantic triples, and semantic similarity, which measures the similarity between two texts. Numerical results show that the proposed algorithm's effectiveness in regards to these two metrics.
\end{itemize}

\begin{figure}[t]
\centering
\includegraphics[width=3in]{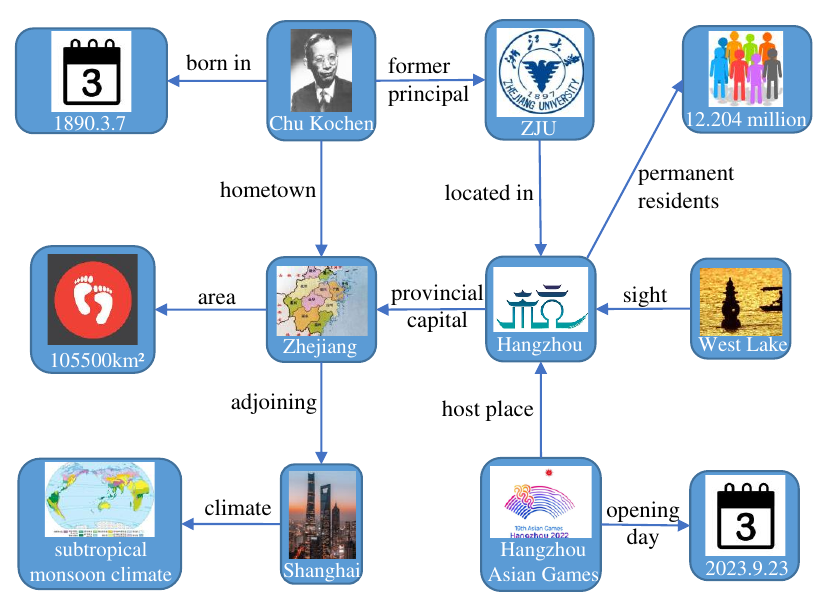}
\vspace{-.5em}
\caption{An example of knowledge graph.}
\label{fig1}
\vspace{-.5em}
\end{figure}

Semantic communication usually relies on semantic information extraction, which can remove redundant information in the original data that is less relevant to the task. There are many methods to extract semantic information, and one of them is knowledge graph. An example of knowledge graph is shown in Fig. \ref{fig1}. Knowledge graph can carry a large amount of information with a small amount of data \cite{zou2020survey}. Therefore, it is a frequently used tool for semantic information extraction \cite{yang2023energy}.

To our knowledge, few research has combined knowledge graph and random theory with wireless communication scenarios. For example, in a wireless resource-constrained scenario, after obtaining a knowledge graph from text data, if there are no sufficient communication resources to transmit the complete knowledge graph within a limited time delay, it is necessary to compress the original knowledge graph. However, there are few such knowledge graph compression methods in the existing research.

We adapt the work in \cite{zhang2022deepke} to generate semantic triples from text data and the confidences of each triple. Then, we propose an algorithm based on probability graph to select a part of the raw triples, which realizes the compression of knowledge graph. 


\section{System Model}
In order to achieve semantic information extraction of text data and reduce the burden of communication networks, this paper uses information extraction techniques in natural language processing (NLP) to extract semantic triples from text data. Secondly, to measure the performance of our algorithm, we propose two metrics called semantic uncertainty and semantic similarity. Those parts will be elaborated in the following separately.

\subsection{Semantic Information Extraction}
This step will first perform named entity recognition on the text data and output the tagged text. Then, the tagged text is fed into the relation extraction model to extract the relations among entities. The final extracted information is represented by a triple (\emph{head, relation, tail}). In performing the relation extraction task, a relation set consisting of several specific relations is available in advance, and the model finally outputs the probability of each specific relation among the entities. Currently, relation extraction still faces some challenges with an accuracy of about 80\% \cite{wang2022deepstruct}. Therefore, the initial triples obtained from text data cannot all be correct, and subsequent steps are needed to minimize the impact of inaccurate triples.

In the studied model, we use $w_n$ to represent a word, a symbol, or a punctuation in the text data. Hereinafter, $w_n$ is called a token. Based on this, the original text data can be represented by an ordered sequence of tokens as shown below:
\vspace{-.5em}
\begin{equation}\label{eq1}\vspace{-.5em}
    L=\{w_1,w_2, \cdots ,w_n, \cdots ,w_N\}, \quad \forall w_n \in\mathcal V,
\end{equation}
where $\mathcal V$ is the vocabulary and $N$ is the number of tokens in $L$. For instance, assuming that the text data need to be transmitted is ``\emph{Apple is a kind of fruit.}''. Hence, in this example, $L=\{[Apple],[is],[a],[kind],[of],[fruit],[.]\}$, where $w_1=[Apple], w_2=[is], w_3=[a], w_4=[kind], w_5=[of], w_6=[fruit], w_7=[.]$.

In the applied model, the semantic information extracted from text data is represented by knowledge graph. A knowledge graph is a structured representation of facts, which consists of entities and relations. The knowledge in a knowledge graph can be represented by a triple (\emph{head, relation, tail}). The knowledge graph consists of a series of nodes and edges \cite{fensel2020introduction}. To be more specific, the node in knowledge graph is called an \emph{entity}, representing the object or concept in the real world. Define entity $j$ in text data $L$ as $e_j$, which consists of a series of tokens in $L$. The edge in knowledge graph represents the \emph{relation} of two entities. Define the relation between entity pair ($e_j,e_k$) as $r_{jk} \in\mathcal R$, where $\mathcal R$ is the relation set.

According to existing NLP techniques, the probability that all the relations in the relation set corresponding to each triple can be obtained. Further, the corresponding entropy, denoted as $h_{jk}$, can be obtained using the following formula:

\vspace{-.5em}
\begin{equation}\label{eq2}\vspace{-.5em}
    h_{jk}=-\sum_{i=1}^A (p_{jk\_i}\log_2 p_{jk\_i}),
\end{equation}
where $A$ is the total number of relations in the relation set $\mathcal R$, and $p_{jk\_i}$ is the confidence of each relation for entity pair ($e_j,e_k$).

The smaller the $h$ is, the more explicit the relations are, and the larger the $h$ is, the vice versa. Therefore, it may be appropriate to expand and write the triples in the form of $(e_j,r_{jk},e_k,h_{jk})$.

\begin{figure}[t]
\centering
\includegraphics[width=3in]{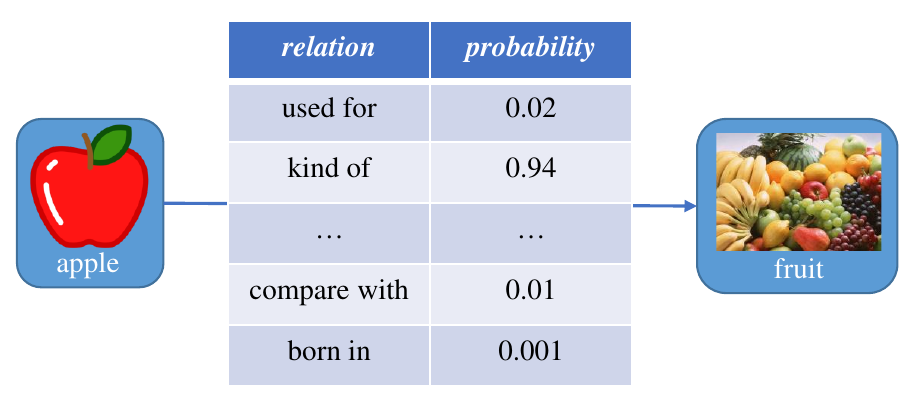}
\vspace{-.5em}
\caption{Each edge between two entities consists of several relation probabilities.}
\vspace{-.5em}
\label{fig2}
\end{figure}

Because the dimension of entropy is added based on probability, the result obtained in this step is not a traditional triple knowledge graph, but a probability graph version of the knowledge graph. The relation between each two nodes is not absolute, but consists of several relations corresponding to probabilities jointly, as shown in Fig. \ref{fig2}.

According to the knowledge graph, the semantic information in the text data $L$ can be written in the following form:

\vspace{-.5em}
\begin{equation}\label{eq3}\vspace{-.5em}
    \mathbb{G}=\{\varepsilon^1,\varepsilon^2,\cdots,\varepsilon^g,\cdots,\varepsilon^{G}\},
\end{equation}
where $\varepsilon^g=(e_j^g,r_{jk}^g,e_k^g,h_{jk}^g\ ), j\neq k$, and $G$ is the total number of quadruples in $\mathbb{G}$. After obtaining the above knowledge graph, we use our algorithm to compress it and extract the relatively important information from it. The extracted important semantic information is represented as:

\vspace{-.5em}
\begin{equation}\label{eq4}\vspace{-.5em}
    \mathbb{G^\prime}=\{\varepsilon^{\prime1},\varepsilon^{\prime2},\cdots,\varepsilon^{\prime h},\cdots,\varepsilon^{\prime H}\},
\end{equation}
where $H$ is the total number of triples after compressing.

\subsection{Semantic Uncertainty}
To evaluate the quality of the semantic information extraction, we proposed a metric called \emph{semantic uncertainty}. Based on the probability, the semantic uncertainty can reflect the ambiguity of the semantic information. We define the semantic uncertainty as:

\vspace{-.5em}
\begin{equation}\label{eq5}\vspace{-.5em}
    SU=\sum_{h \in \mathbb{G^\prime}} h,
\end{equation}
where $h$ is the entropy of each selected quadruple.

As we all know, entropy can measure the uncertainty of an event, the greater the uncertainty, the greater the entropy. Here, $h$ represents the uncertainty of a triple in the knowledge graph, the greater the $h$, the less certain the semantic relation. Therefore, the lower the $SU$, the lower the uncertainty of the semantic information and the better the extraction quality.

\subsection{Semantic Similarity}
Another way to evaluate the quality of semantic extraction is to recover the extracted triples into text and then compare this recovered text with the original text to evaluate the similarity between them. The text recovery can be done with the help of Generative Pre-trained Transformer \cite{floridi2020gpt}.

Thus, we proposed \emph{semantic similarity} to measure the semantic similarity between two texts. The semantic similarity consists of two parts: semantic accuracy and semantic completeness. The semantic accuracy is defined as follows:

\vspace{-.5em}
\begin{equation}\label{eq7}\vspace{-.5em}
    A\left(\mathbb{G}^{\prime}\right)=\frac{\sum_{m=1}^M \min \left\{\sigma\left(L^{\prime}\left(\mathbb{G}^{\prime}\right), e_m\right), \sigma\left(L, e_m\right)\right\}}{\sum_{m=1}^M \sigma\left(L^{\prime}\left(\mathbb{G}^{\prime}\right), e_m\right)},
\end{equation}
where $M$ is the total number of different entities in the extracted triples, $\sigma\left(L^{\prime}\left(\mathbb{G}^{\prime}\right), e_m\right)$ is the number of occurrences of the entity $e_m$ in the recovered text $L^{\prime}\left(\mathbb{G}^{\prime}\right)$, and $\sigma\left(L, e_m\right)$ is the number of occurrences of the entity $e_m$ in the original text $L$. The semantic completeness is defined as follows:

\vspace{-.5em}
\begin{equation}\label{eq8}\vspace{-.5em}
    C\left(\mathbb{G}^{\prime}\right)=\frac{\sum_{m=1}^M \min \left\{\sigma\left(L^{\prime}\left(\mathbb{G}^{\prime}\right), e_m\right), \sigma\left(L, e_m\right)\right\}}{\sum_{m=1}^M \sigma\left(L, e_m\right)}.
\end{equation}

Based on equation (\ref{eq7}) and (\ref{eq8}), the semantic similarity can be written as:

\vspace{-.5em}
\begin{equation}\label{eq9}\vspace{-.5em}
    SS\left(\mathbb{G}^{\prime}\right) = \theta\left(\mathbb{G}^{\prime}\right)\frac{A\left(\mathbb{G}^{\prime}\right)C\left(\mathbb{G}^{\prime}\right)}{\varphi A\left(\mathbb{G^{\prime}}\right)+\left(1-\varphi\right) C\left(\mathbb{G}^{\prime}\right)},
\end{equation}
where $\varphi$ is a parameter that regulates the contribution of semantic accuracy and semantic completeness to semantic similarity, $\varphi \in \left[0,1\right]$. A larger $\varphi$ will increase the impact of semantic accuracy on semantic similarity, and vice versa. $\theta\left(\mathbb{G}^{\prime}\right)$ is defined as:

\vspace{-.5em}
\begin{equation}\label{eq10}\vspace{-.5em}
    \theta\left(\mathbb{G}^{\prime}\right) = \sum_{p \in \mathbb{G}^{\prime}} p,
\end{equation}
where $p$ is the largest relation probability in each extracted triples.

Specifically, for a certain entity $e_m$ corresponding to the recovered text $L^{\prime}\left(\mathbb{G}^{\prime}\right)$, if it appears more times in the recovered text $L^{\prime}\left(\mathbb{G}^{\prime}\right)$ than in the original text $L$, then the semantic accuracy decreases; conversely, the semantic completeness decreases.

\section{Knowledge Graph Important Information Extraction Algorithm}

To further extract the backbone information with high confidence, this paper uses techniques which are related to probability graph to select the best combination.

A piece of text data usually has a central concept around which the rest of the text basically revolves. In knowledge graph, we call the central concept of the corresponding text as initial node. The scheme to determine initial node $e_1$ is as follows: the node with the most occurrences in $\mathbb{G}$ is selected as the initial node; if it is not unique, we choose the node that appears earlier in the text.

Next, the relational distance $d(e_i,e_j)$ is defined as the minimum number of edges to be traversed from $e_i$ to $e_j$, when the directionality of edges is not considered. The relational distance can measure the degree of association between two entities.

\begin{figure}[t]
\centering
\includegraphics[width=3.2in]{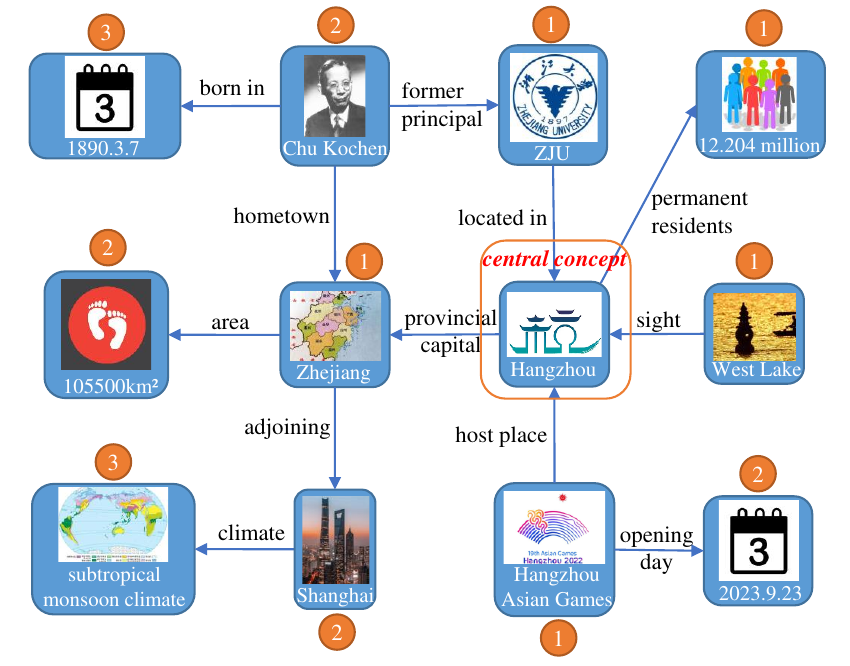}
\vspace{-.5em}
\caption{An example of central concept and relational distance.}
\vspace{-.5em}
\label{fig3}
\end{figure}

For example, in Fig.~\ref{fig3}, there are five edges connected at the node \emph{Hangzhou}, which means node \emph{Hangzhou} appears five times in $\mathbb{G}$. Observing the other nodes, we find that none of them has more occurrences than node \emph{Hangzhou}, so \emph{Hangzhou} is the central concept. The numbers in the orange circles in Fig.~\ref{fig3} represent the relational distance of each node from the initial node \emph{Hangzhou}. For instance, node \emph{West Lake} only needs to pass edge \emph{sight} to reach the initial node \emph{Hangzhou}, so the relational distance of entity \emph{West Lake} is one. Moreover, node \emph{1890.3.7} needs to pass edge \emph{born in}, edge \emph{former principle} and edge \emph{located in}, or edge \emph{born in}, edge \emph{hometown} and edge \emph{provincial capital} to reach the initial node \emph{Hangzhou}, so the relational distance of entity \emph{1890.3.7} is three.

Then, we introduce two parameters:

\begin{itemize}
\item the compression coefficient $K$
\item the maximum depth $D$
\end{itemize}

 The compression coefficient $K$ satisfies $K \in \left(0,1\right]$ and $K=\frac{H}{G}$, where $H$ is the total number of selected quadruples in $\mathbb{G}$. The maximum depth $D$ satisfies $d(e_1,e_i) \leq D$, where $d(e_1,e_i)$ is the relational distance from the initial node $e_1$ to $e_i$, where $e_i \in \mathbb{G}^\prime$. The relational distance $D$ can avoid that the entities in the extracted triple are too far from the central concept.
 Note that in wireless resource constrained scenarios, the number of transmitted bits are limited and the value of $H$ can be modeled as, $Hr=t B\log_2 \left(1+\frac {Ph}{\sigma^2}\right)$, where $r$ is the number of bits for each quadruple in $\mathbb{G}$, $t$ is the communication time, $B$ and $P$ are respectively the limited communication bandwidth and power, $h$ is the channel gain from the transmitter to the receiver, and $\sigma^2$ is the noise power. Here, we consider constant wireless resource allocation, and thus $H$ can be modeled as a constant.

Based on the considered model, the semantic information extraction problem with probability graph can be formulated as:
\begin{subequations}\label{eq6}\vspace{-.5em}
    \begin{align}
        \mathop{\min}_{\mathbb{G}^{\prime}} \quad & \sum_{h \in \mathbb{G}^\prime} h , \tag{\ref{eq6}}\\
        \textrm{s.t.} \quad & d(e_1,e_i) \leq D, \quad \forall e_i \in \mathbb{G}^{\prime}, \label{a1}\\
        \quad & K=\frac{H}{G}. \label{a2}
    \end{align}
\end{subequations}

Note that there is a situation that these two constraints cannot be satisfied at the same time. For instance, $H$ is possible to be larger than the number of quadruples that satisfy constraint (\ref{a1}). In this situation, we appropriately relax the restrictions of constraint (\ref{a1}) by adding the maximum depth $D$. We add one to $D$ at a time, so that the number of quadruples satisfying constraint (\ref{a1}) can increase. We keep doing this until the number of quadruples satisfying constraint (\ref{a1}) is greater than or equal to $H$. At this time, constraint (\ref{a1}) and constraint (\ref{a2}) can be satisfied at the same time.

To solve the above problem, graph-related algorithms are required. First, calculate the relational distance of each node to the initial node based on Floyd's algorithm 
which calculates the complete shortest path, and delete the quadruples with relational distance greater than $D$. Then, we use efficient sorting algorithm to select quadruples with small $h$ among eligible quadruples.

\section{Simulation Results and Analysis}
\begin{figure}[t]
\centering
\includegraphics[width=2.8in]{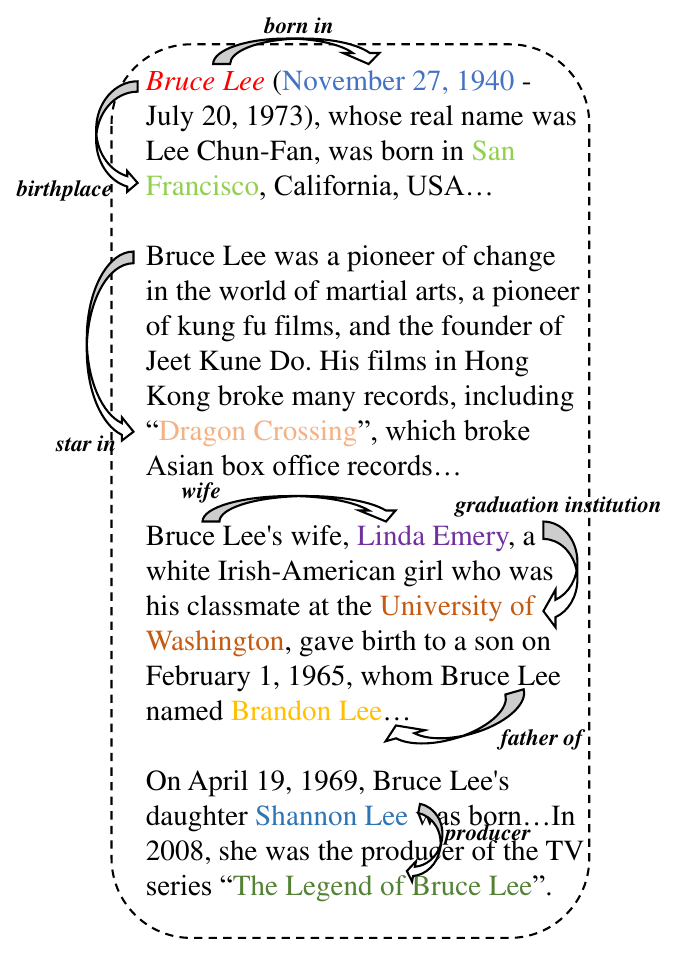}
\vspace{-.5em}
\caption{An example of an original text and the extracted semantic information about ``Bruce Lee''.}
\label{fig4}
\vspace{-.5em}
\end{figure}

We tested our proposed algorithm on a local dataset containing text data from multiple domains such as people, companies, schools, and movies. The texts in the dataset were collected from news or websites like Wikipedia, and each piece of text contains hundreds of words. In Fig. \ref{fig4}, we illustrate part of the original text and part of the extracted semantic information about ``Bruce Lee'' as an example.

\begin{figure}[t]
\centering
\includegraphics[width=3in]{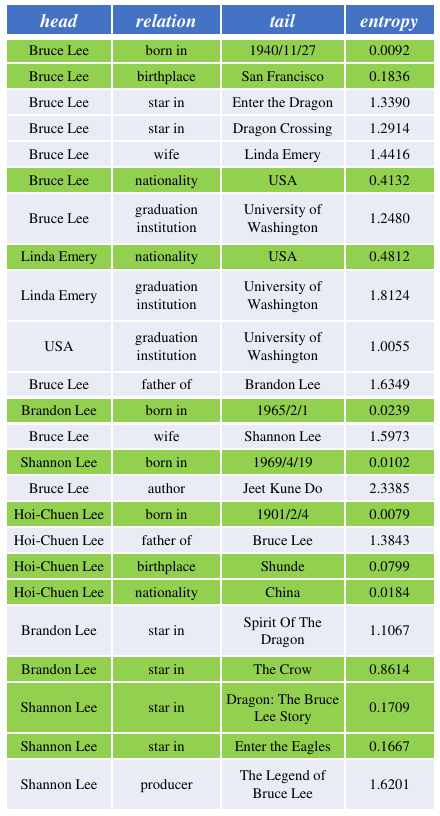}
\vspace{-.5em}
\caption{Original quadruples obtained from the text data about ``Bruce Lee'', and the selected quadruples using our algorithm when $K=0.5$ and $D=2$. Note that the selected quadruples are marked in green.}
\label{fig5}
\vspace{-.5em}
\end{figure}

Using NLP techniques, we can obtain a series of quadruples from the text data, and we take these quadruples as the input of our algorithm. We set $K=0.5$ and $D=2$ to run the simulation, and the result is shown in Fig. \ref{fig5}. As we can see, the selected quadruples have low entropy value, which means they have relatively more explicit semantic relations according to the text.

\begin{figure}[htb]
\centering
\vspace{-.5em}
\includegraphics[width=3in]{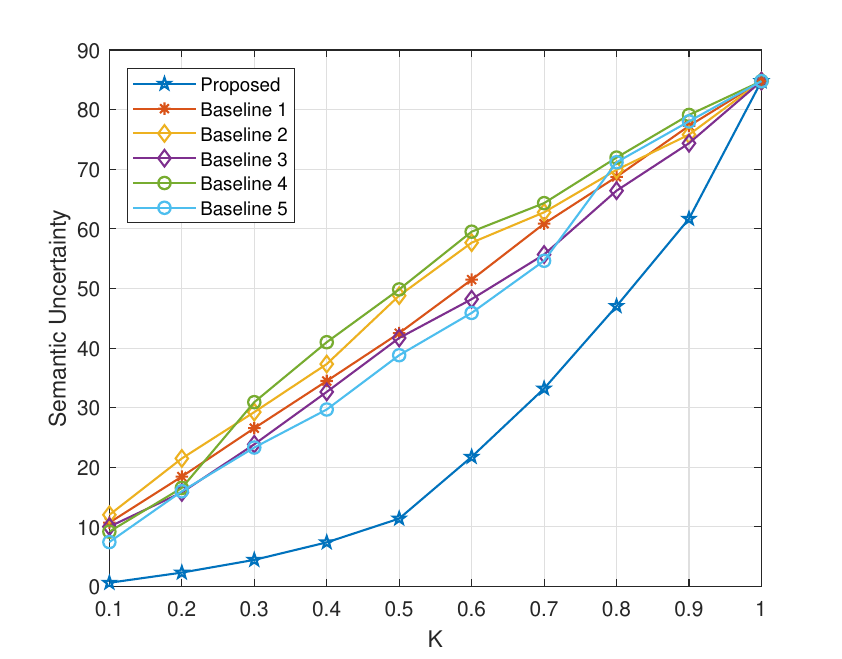}
\caption{The relationship between the compression coefficient $K$ and the semantic uncertainty $SU$ when the maximum depth $D=2$. The result of baseline 1 is the average of 100 runs.}
\vspace{-.5em}
\label{fig6}
\end{figure}

As mentioned above, the compression coefficient $K$ determines the compression ratio of the original knowledge graph. In order to find out the relationship between the compression coefficient $K$ and the semantic uncertainty $SU$, we changed $K$ from 0.1 to 1 and calculated $SU$ respectively. For comparison, we considered several baselines: 1) chooses quadruples randomly, 2) extracts by the number of occurrences of the entity from most to least, 3) extracts by the number of occurrences of the entity from least to most, 4) extracts the triples in order of appearance from front to back, and 5) extracts the triples in order of appearance from back to front. The result is shown in Fig. \ref{fig6}.

In Fig. \ref{fig6}, we can see that, in our proposed algorithm, as the compression coefficient $K$ increases, the semantic uncertainty $SU$ increases. This is because as the compression coefficient $K$ increases, the total number of selected quadruples increases, and every quadruple will bring additional semantic uncertainty to $SU$. Fig. \ref{fig6} also shows that the proposed algorithm always has lower $SU$ than baseline algorithms, except when $K=1$. In the case of $K=1$, all original quadruples are chosen, so the $SU$ value of all algorithms are the same. The result shows that our proposed algorithm can extract more explicit information from the original knowledge graph.

\begin{figure}[htb]
\centering
\vspace{-.5em}
\includegraphics[width=3.2in]{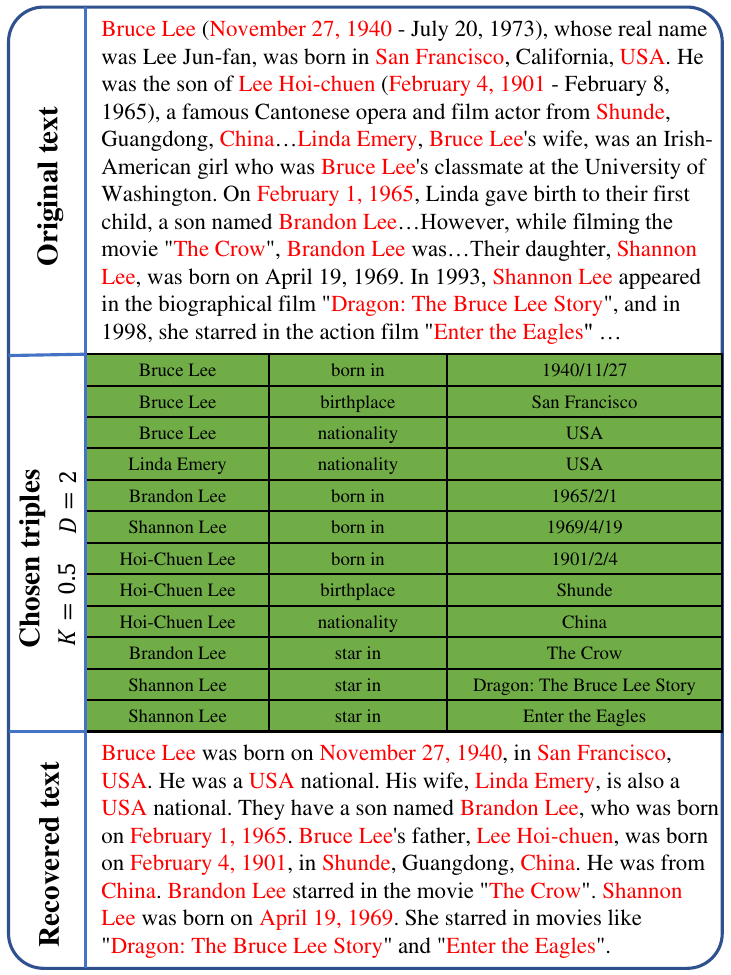}
\caption{An example of the original text, extracted triples, and corresponding recovered text. The entities in chosen triples are marked in red.}
\vspace{-.5em}
\label{fig8}
\end{figure}

\begin{figure}[htb]
\centering
\includegraphics[width=3in]{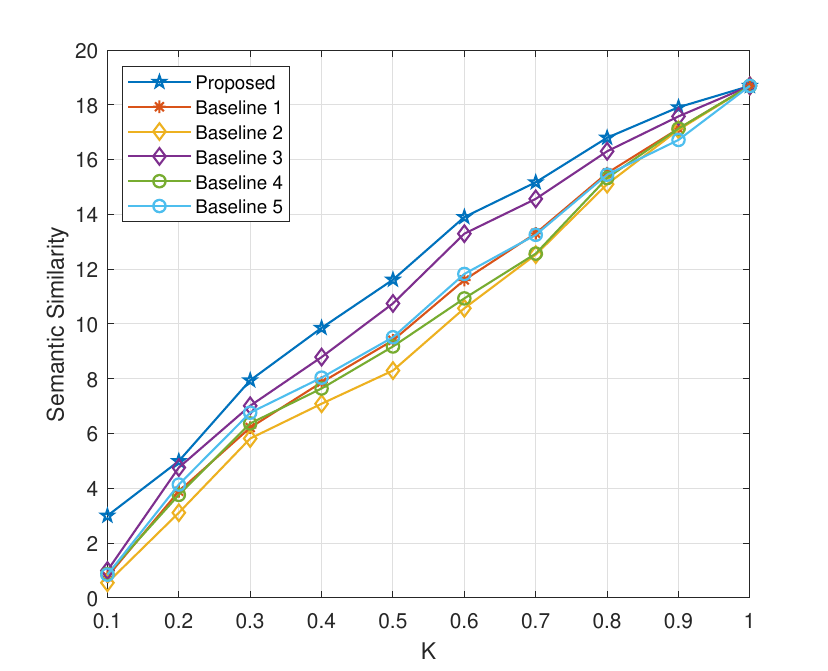}
\vspace{-.5em}
\caption{The relationship between the compression coefficient $K$ and the semantic similarity $SS$ when the maximum depth $D=2$.}
\vspace{-1em}
\label{fig9}
\end{figure}

Further, the quality of semantic extraction is evaluated by semantic similarity after recovering the extracted triples into text. An example of the original text, extracted triples, and corresponding recovered text is presented in Fig. \ref{fig8}. The result in Fig. \ref{fig9} demonstrates that an increase in the compression coefficient $K$ leads to an increase in semantic similarity $SS$. This is because the selection of triples for text recovery increases with higher compression coefficients, resulting in a greater similarity between the recovered text and the original text. The front segment of the proposed algorithm's semantic similarity increases more rapidly than the end segment because the triples used in the former are of higher semantic quality and provide more accurate semantic information. Although the gap between the proposed algorithm and other algorithms decreases after the triples are recovered into text, the proposed algorithm still retains some advantages.

\section{Conclusion}
In this paper, we leveraged knowledge graphs to model the semantic information of textual data and presented a novel algorithm for extracting important information from these graphs. To measure the performance of our approach, we introduced two metrics: semantic uncertainty, which reflects the clarity of the semantic triples, and semantic similarity, which measures the similarity between the recovered text and the original text. Our approach introduced the concept of relational probability into knowledge graphs and extends traditional triples into quadruples. We also proposed two parameters, the compression coefficient $K$ and the maximum depth $D$, which determine the number of selected quadruples and the compactness of the extracted information, respectively. We then formulated an optimization problem that aims to minimize the information entropy of the selected semantic information while satisfying the constraints of the compression coefficient and maximum depth. Our simulation results demonstrate the effectiveness of the proposed knowledge graph important information extraction algorithm.

With the ability to effectively reduce the amount of data that needs to be transmitted while preserving the optimal textual semantics, our algorithm can play an important role when the communication resources are severely limited. This capability becomes especially valuable in satellite communications, underwater sensor networks and other contexts with constrained communication resources.

Looking ahead, an intriguing and challenging research direction is the joint communication and computation resource allocation while incorporating probability graph of semantic information extraction.

\bibliographystyle{IEEEtran}
\bibliography{IEEEabrv,SIETDPA}

\begin{thebibliography}{10}
\providecommand{\url}[1]{#1}
\csname url@samestyle\endcsname
\providecommand{\newblock}{\relax}
\providecommand{\bibinfo}[2]{#2}
\providecommand{\BIBentrySTDinterwordspacing}{\spaceskip=0pt\relax}
\providecommand{\BIBentryALTinterwordstretchfactor}{4}
\providecommand{\BIBentryALTinterwordspacing}{\spaceskip=\fontdimen2\font plus
\BIBentryALTinterwordstretchfactor\fontdimen3\font minus
  \fontdimen4\font\relax}
\providecommand{\BIBforeignlanguage}[2]{{%
\expandafter\ifx\csname l@#1\endcsname\relax
\typeout{** WARNING: IEEEtran.bst: No hyphenation pattern has been}%
\typeout{** loaded for the language `#1'. Using the pattern for}%
\typeout{** the default language instead.}%
\else
\language=\csname l@#1\endcsname
\fi
#2}}
\providecommand{\BIBdecl}{\relax}
\BIBdecl

\bibitem{shannon1948mathematical}
C.~E. Shannon, ``A mathematical theory of communication,'' \emph{The Bell
  system technical journal}, vol.~27, no.~3, pp. 379--423, 1948.

\bibitem{gunduz2022beyond}
D.~G{\"u}nd{\"u}z, Z.~Qin, I.~E. Aguerri, H.~S. Dhillon, Z.~Yang, A.~Yener,
  K.~K. Wong, and C.-B. Chae, ``Beyond transmitting bits: Context, semantics,
  and task-oriented communications,'' \emph{IEEE Journal on Selected Areas in
  Communications}, vol.~41, no.~1, pp. 5--41, 2022.

\bibitem{2023big}
Z.~Chen, Z.~Zhang, and Z.~Yang, ``Big {AI} models for 6g wireless networks:
  Opportunities, challenges, and research directions,'' \emph{arXiv preprint
  arXiv:2308.06250}, 2023.

\bibitem{chaccour2022less}
C.~Chaccour, W.~Saad, M.~Debbah, Z.~Han, and H.~V. Poor, ``Less data, more
  knowledge: Building next generation semantic communication networks,''
  \emph{arXiv preprint arXiv:2211.14343}, 2022.

\bibitem{9832831}
Y.~Wang, M.~Chen, T.~Luo, W.~Saad, D.~Niyato, H.~V. Poor, and S.~Cui,
  ``Performance optimization for semantic communications: An attention-based
  reinforcement learning approach,'' \emph{IEEE Journal on Selected Areas in
  Communications}, vol.~40, no.~9, pp. 2598--2613, 2022.

\bibitem{han2022semantic}
T.~Han, Q.~Yang, Z.~Shi, S.~He, and Z.~Zhang, ``Semantic-preserved
  communication system for highly efficient speech transmission,'' \emph{IEEE
  Journal on Selected Areas in Communications}, vol.~41, no.~1, pp. 245--259,
  2022.

\bibitem{10024766}
W.~Xu, Z.~Yang, D.~W.~K. Ng, M.~Levorato, Y.~C. Eldar, and M.~Debbah, ``Edge
  learning for {B5G} networks with distributed signal processing: Semantic
  communication, edge computing, and wireless sensing,'' \emph{IEEE J. Sel.
  Topics Signal Process.}, vol.~17, no.~1, pp. 9--39, Jan. 2023.

\bibitem{peng2022robust}
X.~Peng, Z.~Qin, D.~Huang, X.~Tao, J.~Lu, G.~Liu, and C.~Pan, ``A robust deep
  learning enabled semantic communication system for text,'' in \emph{GLOBECOM
  2022-2022 IEEE Global Communications Conference}.\hskip 1em plus 0.5em minus
  0.4em\relax IEEE, 2022, pp. 2704--2709.

\bibitem{xie2021deep}
H.~Xie, Z.~Qin, G.~Y. Li, and B.-H. Juang, ``Deep learning enabled semantic
  communication systems,'' \emph{IEEE Transactions on Signal Processing},
  vol.~69, pp. 2663--2675, 2021.

\bibitem{weng2021semantic}
Z.~Weng and Z.~Qin, ``Semantic communication systems for speech transmission,''
  \emph{IEEE Journal on Selected Areas in Communications}, vol.~39, no.~8, pp.
  2434--2444, 2021.

\bibitem{tong2021federated}
H.~Tong, Z.~Yang, S.~Wang, Y.~Hu, W.~Saad, and C.~Yin, ``Federated learning
  based audio semantic communication over wireless networks,'' in \emph{2021
  IEEE Global Communications Conference (GLOBECOM)}.\hskip 1em plus 0.5em minus
  0.4em\relax IEEE, 2021, pp. 1--6.

\bibitem{xie2020lite}
H.~Xie and Z.~Qin, ``A lite distributed semantic communication system for
  internet of things,'' \emph{IEEE Journal on Selected Areas in
  Communications}, vol.~39, no.~1, pp. 142--153, 2020.

\bibitem{zou2020survey}
X.~Zou, ``A survey on application of knowledge graph,'' in \emph{Journal of
  Physics: Conference Series}, vol. 1487, no.~1.\hskip 1em plus 0.5em minus
  0.4em\relax IOP Publishing, 2020, p. 012016.

\bibitem{yang2023energy}
Z.~Yang, M.~Chen, Z.~Zhang, and C.~Huang, ``Energy efficient semantic
  communication over wireless networks with rate splitting,'' \emph{IEEE J.
  Sel. Areas Commun.}, vol.~41, no.~5, pp. 1484--1495, 2023.

\bibitem{zhang2022deepke}
N.~Zhang, X.~Xu, L.~Tao, H.~Yu, H.~Ye, S.~Qiao, X.~Xie, X.~Chen, Z.~Li, L.~Li
  \emph{et~al.}, ``Deepke: A deep learning based knowledge extraction toolkit
  for knowledge base population,'' \emph{arXiv preprint arXiv:2201.03335},
  2022.

\bibitem{wang2022deepstruct}
C.~Wang, X.~Liu, Z.~Chen, H.~Hong, J.~Tang, and D.~Song, ``Deepstruct:
  Pretraining of language models for structure prediction,'' \emph{arXiv
  preprint arXiv:2205.10475}, 2022.

\bibitem{fensel2020introduction}
D.~Fensel, U.~{\c{S}}im{\c{s}}ek, K.~Angele, E.~Huaman, E.~K{\"a}rle,
  O.~Panasiuk, I.~Toma, J.~Umbrich, A.~Wahler, D.~Fensel \emph{et~al.},
  ``Introduction: what is a knowledge graph?'' \emph{Knowledge graphs:
  Methodology, tools and selected use cases}, pp. 1--10, 2020.

\bibitem{floridi2020gpt}
L.~Floridi and M.~Chiriatti, ``Gpt-3: Its nature, scope, limits, and
  consequences,'' \emph{Minds and Machines}, vol.~30, pp. 681--694, 2020.

\end{thebibliography}

\end{document}